\def\year{2021}\relax
\documentclass[letterpaper]{article} %
\usepackage{aaai21}  %
\usepackage{times}  %
\usepackage{helvet} %
\usepackage{courier}  %
\usepackage[hyphens]{url}  %
\usepackage{graphicx} %
\usepackage{natbib}  %
\usepackage{caption} %
\usepackage{amssymb}
\usepackage{amsmath}
\usepackage{subcaption}
\usepackage{forest}
\usepackage[ruled,vlined,linesnumbered]{algorithm2e}
\usepackage{tabularx}
\usepackage[switch]{lineno}

\definecolor{darkred}{rgb}{0.8, 0.0, 0.0}

\definecolor{purple}{rgb}{0.5,0.0,0.5}

\title{Successor Feature Sets: Generalizing Successor Representations Across Policies}
\author {

        Kiant\'e Brantley,\textsuperscript{\rm 1}
        Soroush Mehri, \textsuperscript{\rm 2}
        Geoffrey J. Gordon \textsuperscript{\rm 2} \\
}
\affiliations {
    \textsuperscript{\rm 1} University of Maryland College Park \\
    \textsuperscript{\rm 2} Microsoft Research \\
    kdbrant@umd.edu, somehri@microsoft.com, ggordon@microsoft.com
}

\begin{document}
\maketitle

\begin{abstract}
Successor-style representations have many advantages for reinforcement learning: for example, they can help an agent generalize from past experience to new goals, and they have been proposed as explanations of behavioral and neural data from human and animal learners. They also form a natural bridge between model-based and model-free RL methods: like the former they make predictions about future experiences, and like the latter they allow efficient prediction of total discounted rewards. However, successor-style representations are not optimized to generalize across policies: typically, we maintain a limited-length list of policies, and share information among them by representation learning or GPI\@. Successor-style representations also typically make no provision for gathering information or reasoning about latent variables.
To address these limitations, we bring together ideas from predictive state representations, belief space value iteration, successor features, and convex analysis: we develop a new, general successor-style representation, together with a Bellman equation that connects multiple sources of information within this representation, including different latent states, policies, and reward functions.
The new representation is highly expressive: for example, it lets us efficiently read off an optimal policy for a new reward function, or a policy that imitates a new demonstration.
For this paper, we focus on exact computation of the new representation in small, known environments, since even this restricted setting offers plenty of interesting questions. 
Our implementation does not scale to large, unknown environments --- nor would we expect it to, since it generalizes POMDP value iteration, which is difficult to scale. However, we believe that future work will allow us to extend our ideas to approximate reasoning in large, unknown environments. We conduct experiments to explore which of the potential barriers to scaling are most pressing.
\end{abstract}

\section{Introduction}

\noindent We describe a new representation for decision-theoretic planning, reinforcement learning, and imitation learning: the \emph{successor feature set}. This representation generalizes a number of previous ideas in the literature, including successor features and POMDP/PSR value functions. 
Comparing to these previous representations: successor features assume a fixed policy or list of policies, while our goal is to reason efficiently about many policies at once; value functions assume a fixed reward function, while our goal is to reason efficiently about many reward functions at once.

Roughly, the successor feature set tells us how features of our future observations and actions depend on our current state and our choice of policy. 
More specifically, the successor feature set is a convex set of matrices; each matrix corresponds to a policy $\pi$, and describes how the features we will observe in the future depend on the current state under $\pi$.

The successor feature set provides a number of useful capabilities. These include reading off the optimal value function or policy for a new reward function, predicting the range of outcomes that we can achieve starting from a given state, and reading off a policy that imitates a desired state-action visitation distribution. %

We describe a convergent dynamic programming algorithm for computing the successor feature set, generalizing the value iteration algorithm for POMDPs or PSRs. We also give algorithms for reading off the above-mentioned optimal policies and feature-matching policies from the successor feature set. Since the exact dynamic programming algorithm can be prohibitively expensive, we also experiment with randomized numerical approximations.

In this paper we focus on model-based reasoning about successor feature sets \textemdash\ that is, we assume access to an accurate world model. We also focus on algorithms that are exact in the limit of increasing computation. Successor-style representations are of course also extremely useful for approximate reasoning about large, unknown environments, and we believe that many of the ideas discussed here can inform that case as well, but we leave that direction for future work.

To summarize, our contributions are: a new successor-style representation that allows information to flow among different states, policies, and reward functions; algorithms for working with this new representation in small, known environments, including a convergent dynamic programming algorithm and ways to read off optimal policies and feature-matching policies; and computational experiments that evaluate the strengths and limitations of our new representation and algorithms.

\section{Background and Notation}

Our environment is a controlled dynamical system. We interact with it in a sequence of time steps; at each step, all relevant information is encoded in a state vector. Given this state vector, we choose an action. Based on the action and the current state, the environment changes to a new state, emits an observation, and moves to the next time step. We can describe such a system using one of a few related models: a Markov decision process (MDP), a partially-observable Markov decision process (POMDP), or a (transformed) predictive state representation (PSR). We describe these models below, and summarize our notation in Table~\ref{tbl:notation}.

\subsection{MDPs}

An MDP is the simplest model: there are $k$ possible discrete states, numbered $1\ldots k$. The environment starts in one of these states, $s_1$. For each possible action $a\in \{1\ldots A\}$, the \emph{transition matrix} $T_a\in\mathbb R^{k\times k}$ tells us how our state changes if we execute action $a$: $[T_a]_{ij}$ is the probability that the next state is $s_{t+1}=i$ if the current state is $s_t=j$. 

More compactly, we can associate each state $1, 2, \ldots, k$ with a corresponding standard basis vector $e_1, e_2, \ldots, e_k$, and write $q_t$ for the vector at time $t$. (So, if $s_t=i$ then $q_t = e_i$.) Then, $T_a q_t$ is the probability distribution over next states: 
$$P(s_{t+1} \mid q_t, \mathrm{do}~a) = \mathbb E(q_{t+1}\mid q_t, \mathrm{do}~a) = T_a q_t$$
Here we have written $\mathrm{do}~a$ to indicate that choosing an action is an \emph{intervention}.

\subsection{POMDPs}

In an MDP, we get to know the exact state at each time step: $q_t$ is always a standard basis vector. By contrast, in a POMDP, we only receive partial information about the underlying state: at each time step, after choosing our action $a_t$, we see an observation $o_t\in \{1\ldots O\}$ according to a distribution that depends on the next state $s_{t+1}$. The \emph{observation matrix} $D\in\mathbb R^{O\times k}$ tells us the probabilities: $D_{ij}$ is the probability of receiving observation $o_t=i$ if the next state is $s_{t+1} = j$. 

To represent this partial information about state, we can let the state vector $q_t$ range over the probability simplex instead of just the standard basis vectors: $[q_t]_i$ tells us the probability that the state is $s_t=i$, given all actions and observations so far, up to and including $a_{t-1}$ and $o_{t-1}$. The vector $q_t$ is called our \emph{belief state}; we start in belief state $q_1$.

Just as in an MDP, we have $\mathbb E(q_{t+1}\mid q_t,\mathrm{do}~a) = T_a q_t$. But now, instead of immediately resolving $q_{t+1}$ to one of the corners of the simplex, we can only take into account partial state information: if $o_{t}=o$ then by Bayes rule
\begin{eqnarray*}
[q_{t+1}]_i &=& P(s_{t+1}=i\mid q_t, \mathrm{do}~a, o)\\
&=& \frac{P(o\mid s_{t+1}=i)P(s_{t+1}=i\mid q_t, \mathrm{do}~a)}{P(o\mid q_t, \mathrm{do}~a)}\\
&=& \textstyle D_{oi}[T_a q_t]_i\,/\,\sum_{o'} D_{o'i}[T_a q_t]_i
\end{eqnarray*}
More compactly, if $u\in\mathbb R^k$ is the vector of all $1$s, and
$$T_{ao} = \mathrm{diag}(D_{o,\cdot}) T_a$$
where $\mathrm{diag}(\cdot)$ constructs a diagonal matrix from a vector, then our next belief state is
$$q_{t+1} = T_{ao}q_t \,/\, u^T T_{ao}q_t$$
A POMDP is strictly more general than an MDP: if our observation $o_t$ tells us complete information about our next state $s_{t+1}$, then our belief state $q_{t+1}$ will be a standard basis vector. This happens precisely when $[P(o_{t}=i\mid s_{t+1}=j) = 1] \Leftrightarrow [i=j]$.

\subsection{PSRs}

A PSR further generalizes a POMDP: we can think of a PSR as dropping the interpretation of $q_t$ as a belief state, and keeping only the mathematical form of the state update. That is, we no longer require our model parameters to have any interpretation in terms of probabilities of partially observable states; we only require them to produce valid observation probability estimates. (It is possible to interpret PSR states and parameters in terms of experiments called \emph{tests}; for completeness we describe this interpretation in the supplementary material, available online.) %

\begin{table}[t]
\centering
\begin{tabular}{l|l|p{1.75in}}
\textbf{Symbol} & \textbf{Type} & \textbf{Meaning} \\
\hline
$d$ & $\mathbb N$ & dimension of feature vector \\
$k$ & $\mathbb N$ & dimension of state vector \\
$A, O$ & $\mathbb N$ & number of actions, observations \\
$f(q,a)$ & $\mathbb R^d$ & one-step feature function \\
$F_a$ & $\mathbb R^{d\times k}$ & implements $f$: $f(q,a) = F_a q$ \\
$T_{ao}$ & $\mathbb R^{k\times k}$ & transition operator for action, observation \\
$\phi^\pi, \phi^\pi(q)$ & $\mathbb R^{d}$ & successor features for $\pi$ (in state $q$) \\
$A^\pi$ & $\mathbb R^{d\times k}$ & implements $\phi^\pi$: $\phi^\pi(q)= A^\pi q$ \\
$\Phi$ & $\{\mathbb R^{d\times k}\}$ & successor set \\
$\Phi_a, \Phi_{ao}$ & $\{\mathbb R^{d\times k}\}$ & backups of $\Phi$ for actions and observations
\end{tabular}
\caption{Notation quick reference}
\label{tbl:notation}
\end{table}

In more detail, we are given a starting state vector $q_1$, matrices $T_{ao}\in\mathbb R^{k\times k}$, and a normalization vector $u\in\mathbb R^k$. We define our state vector by the recursion
$$q_{t+1} = T_{a_to_t}q_t / u^T T_{a_to_t}q_t$$
and our observation probabilities as
$$P(o_t = o \mid q_t, \mathrm{do}~a) = u^T T_{ao}q_t$$
The only requirement on the parameters is that the observation probabilities $u^T T_{ao}q_t$ should always be nonnegative and sum to 1: under any sequence of actions and observations, if $q_t$ is the resulting sequence of states,
$$(\forall a, o, t) \, u^T T_{ao}q_t \geq 0 \quad 
\textstyle (\forall a, t)\, \sum_o u^T T_{ao}q_t = 1$$
It is clear that a PSR generalizes a POMDP, and therefore also an MDP: we can always take $u$ to be the vector of all $1$s, and set $T_{ao}$ according to the POMDP transition and observation probabilities, so that 
$$[T_{ao}]_{ij} = P(o_t=o, s_{t+1}=i \mid s_t = j, a_t = a)$$
It turns out that PSRs are a \emph{strict} generalization of POMDPs: there exist PSRs whose dynamical systems cannot be described by any finite POMDP\@. An example is the so-called \emph{probability clock}~\cite{jaeger00ooms}.

\subsection{Policy Trees}

\begin{figure}[ht]
\centering
\begin{forest}
for tree={circle,draw, l sep=20pt}
[$\uparrow$, 
    [$\uparrow$,edge label={node[midway,fill=white] {R}} 
      [$\uparrow$,edge label={node[midway,fill=white] {R}} ] 
      [$\downarrow$,edge label={node[midway,fill=white] {G}} ] 
      [$\downarrow$,edge label={node[midway,fill=white] {B}} ]
    ]
    [$\downarrow$,edge label={node[midway,fill=white] {G}} 
      [$\downarrow$,edge label={node[midway,fill=white] {R}} ] 
      [$\downarrow$,edge label={node[midway,fill=white] {G}} ] 
      [$\downarrow$,edge label={node[midway,fill=white] {B}} ]
  ] 
  [$\uparrow$,edge label={node[midway,fill=white] {B}} 
      [$\uparrow$,edge label={node[midway,fill=white] {R}} ] 
      [$\uparrow$,edge label={node[midway,fill=white] {G}} ] 
      [$\downarrow$,edge label={node[midway,fill=white] {B}} ]
  ] 
]
\end{forest}
\caption{An example of a policy tree with actions $\uparrow, \downarrow$ and observations $R, G, B$.}
\label{fig:policy_tree}
\end{figure}
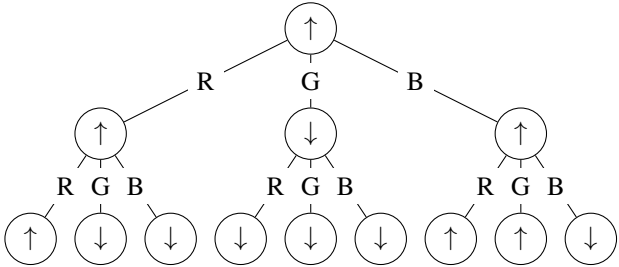

We will need to work with policies for MDPs, POMDPs, and PSRs, handling different horizons as well as partial observability. For this reason, we will use a general policy representation: we will view a policy as a mixture of trees, with each tree representing a deterministic, nonstationary policy. A policy tree's nodes are labeled with actions, and its edges are labeled with observations (Fig.~\ref{fig:policy_tree}).
To execute a policy tree $\pi$, we execute $\pi$'s root action; then, based on the resulting observation $o$, we follow the edge labeled $o$ from the root, leading to a subtree that we will call $\pi(o)$. To execute a mixture, we randomize over its elements. If desired we can randomize lazily, committing to each decision just before it affects our actions. We will work with finite, balanced trees, with depth equal to a horizon $H$; we can reason about infinite-horizon policies by taking a limit as $H\to\infty$.

\section{Imitation by Feature Matching}
\label{sec:feature-matching}

Successor feature sets have many uses, but we will start by motivating them with the goal of imitation.
Often we are given demonstrations of some desired behavior in a dynamical system, and we would like to imitate that behavior. There are lots of ways to specify this problem, but one reasonable one is \emph{apprenticeship learning} \citep{abbeel2004apprenticeship} or \emph{feature matching}. In this method, we define features of states and actions, and ask our learner to match some statistics of the observed features of our demonstrations.

In more detail, given an MDP, define a vector of features of the current state and action, $f(s,a)\in\mathbb R^d$; we call this the \emph{one-step} or \emph{immediate} feature vector. We can calculate the observed discounted features of a demonstration: if we visit states and actions $s_1, a_1, s_2, a_2, s_3, a_3, \ldots$, then the empirical discounted feature vector is
$$f(s_1,a_1) + \gamma f(s_2,a_2) + \gamma^2 f(s_3, a_3) + \ldots $$
where $\gamma\in[0,1)$ is our \emph{discount factor}.
We can average the feature vectors for all of our demonstrations to get a \emph{demonstration} or \emph{target} feature vector $\phi^d$. 

Analogously, for a policy $\pi$, we can define the \emph{expected} discounted feature vector:
$$\phi^\pi = \mathbb E_\pi \left[ \sum_{t=1}^\infty \gamma^{t-1} f(s_t, a_t)\right]$$
We can use a finite horizon $H$ by replacing $\sum_{t=1}^\infty$ with $\sum_{t=1}^H$ in the definitions of $\phi^d$ and $\phi^\pi$; in this case we have the option of setting $\gamma=1$.

\begin{figure}[t]
\centering
\begin{subfigure}{0.4\columnwidth}
\includegraphics[width=\linewidth]{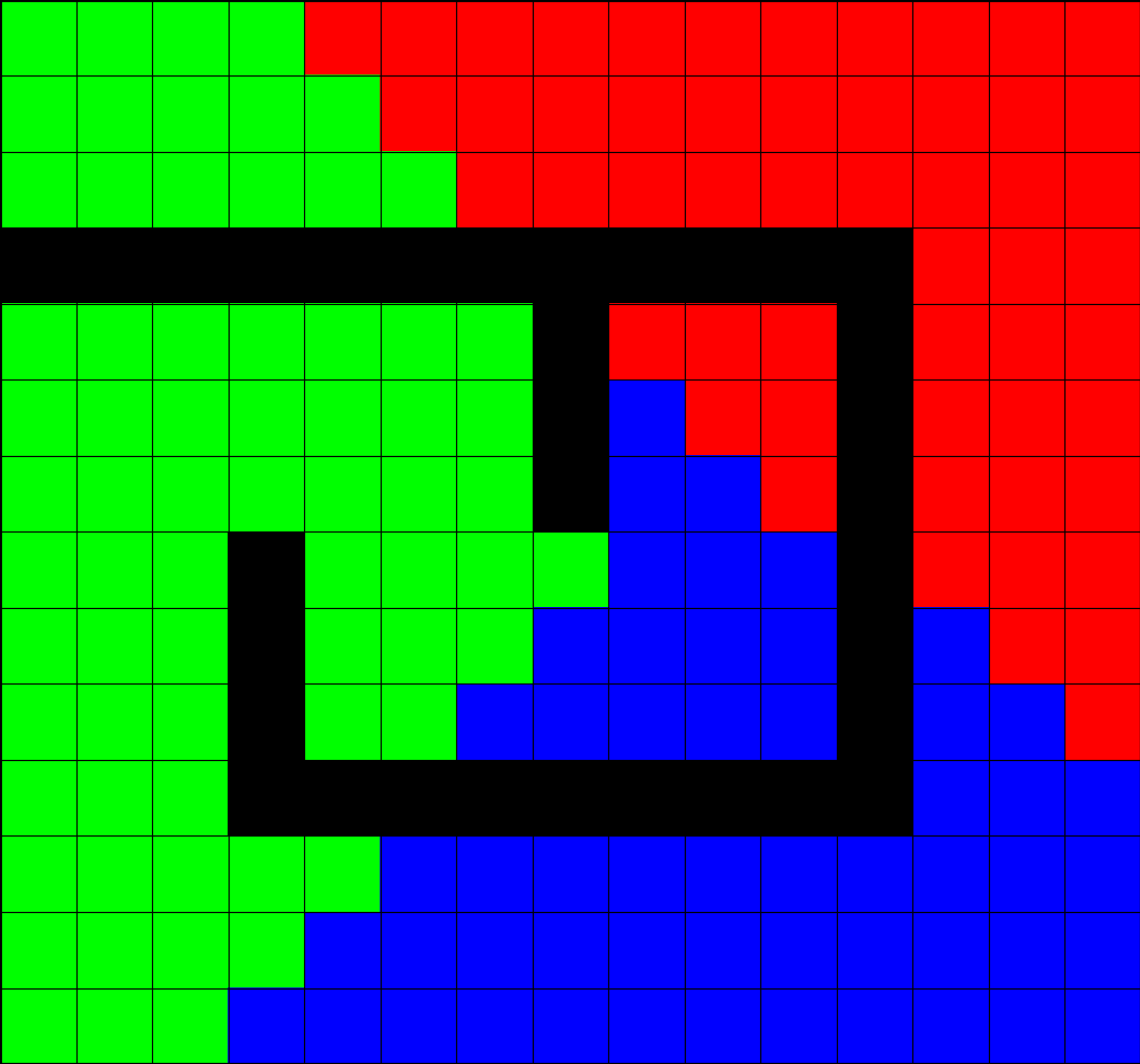}
\caption{$f$ of maze MDP.}
\label{fig:sr_f_maze}
\end{subfigure}
\qquad
\begin{subfigure}{0.4\columnwidth}
\includegraphics[width=\linewidth]{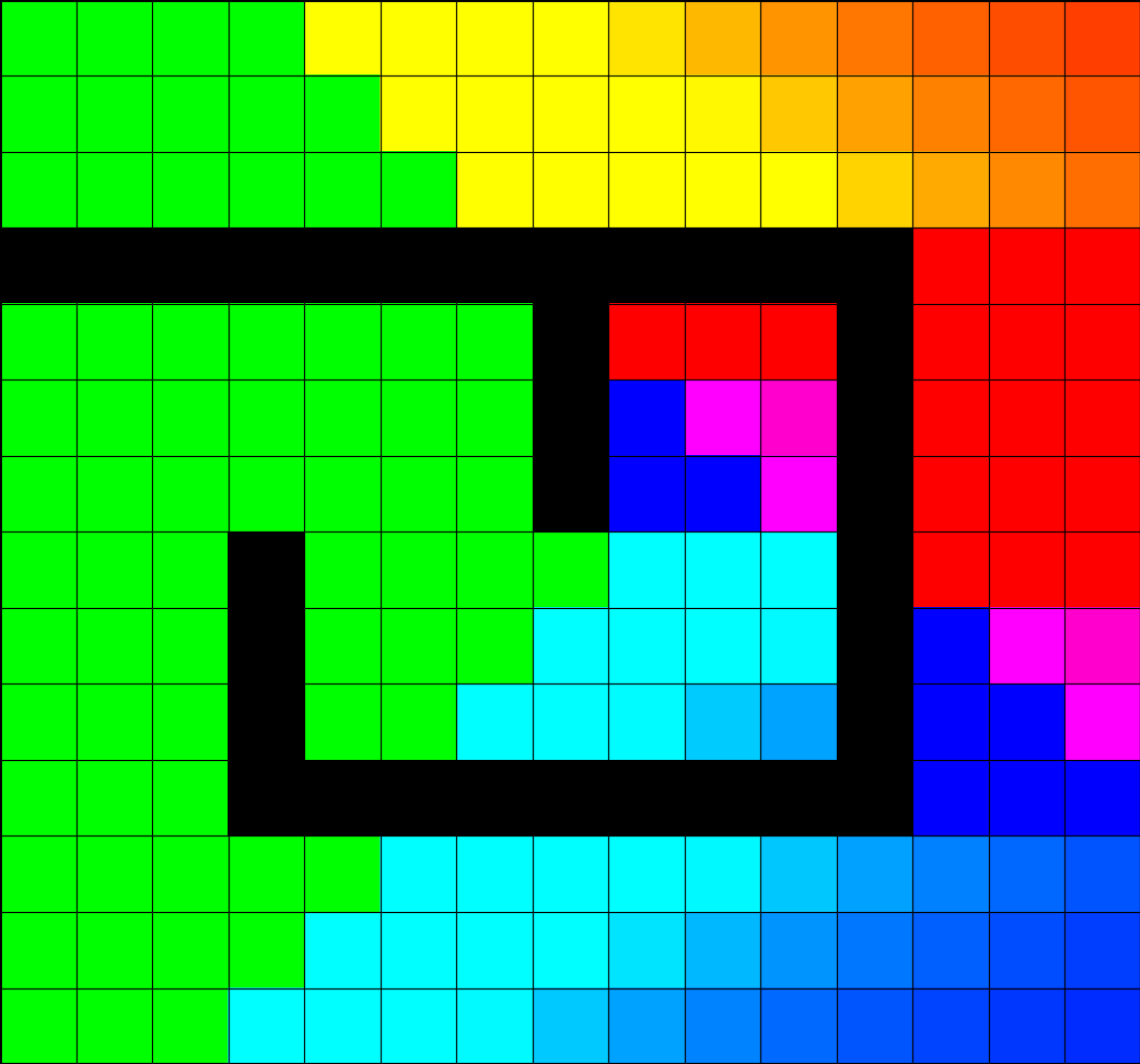}
\caption{$\phi^{\text{go-left}}$ with $\gamma=0.75$.}
\label{fig:sr_phi_maze}
\end{subfigure}

\caption{Maze environment example.}
\label{fig:sr_maze}
\end{figure}

Given a target feature vector in any of these models, we can ask our learner to design a policy that matches the target feature vector in expectation. That is, we ask the learner to find a policy $\pi$ with
$$\phi^\pi = \phi^d$$
For example, suppose our world is a simple maze MDP like Fig.~\ref{fig:sr_f_maze}. Suppose that our one-step feature vector $f(s,a)\in [0,1]^3$ is the RGB color of the current state in this figure, and that our discount is $\gamma=0.75$. If our demonstrations spend most of their time toward the left-hand side of the state space, then our target vector will be something like $\phi^d = [0.5, 3, 0.5]^T$: the green feature will have the highest expected discounted value. On the other hand, if our demonstrations spend most of their time toward the bottom-right corner, we might see something like $\phi^d=[2,1,1]^T$, with the blue feature highest.

\section{Successor Features}

To reason about feature matching, it will be important to predict how the features we see in the future depend on our current state. To this end, we define an analog of $\phi^\pi$ where we vary our start state, called the \emph{successor feature representation} \cite{dayan1993improving,barreto2017successor}:
$$\phi^\pi(s) = \mathbb E_\pi\left[\sum_{t=1}^\infty \gamma^{t-1} f(s_t, a_t)\,\bigg|\, \mathrm{do}~s_1 = s\right]$$
This function associates a vector of expected discounted features to each possible start state. We can think of $\phi^\pi(\cdot)$ as a generalization of a value function: instead of predicting total discounted rewards, it predicts total discounted feature vectors. In fact, the generalization is strict: $\phi^\pi(\cdot)$ contains enough information to compute the value function for any one-step reward function of the form $r^Tf(s,a)$, via $V^\pi(s)=r^T\phi^\pi(s)$.

For example, in Fig.~\ref{fig:sr_phi_maze}, our policy is to always move left. The corresponding successor feature function looks similar to the immediate feature function, except that colors will be smeared rightward. The smearing will stop at walls, since an agent attempting to move through a wall will stop.

\section{Extension to POMDPs and PSRs}

We can generalize the above definitions to models with partial observability as well. This is not a typical use of successor features: reasoning about partial observability requires a model, while successor-style representations are often used in model-free RL\@. However, as \citet{lehnert19lsfm} point out, the state of a PSR is already a prediction about the future, so incorporating successor features into these models makes sense.

In a POMDP, we have a belief state $q\in\mathbb R^k$ instead of a fully-observed state. We define the immediate features of $q$ to be the expected features of the latent state: 
$$f(q, a) = \sum_{s=1}^k q(s) f(s, a)$$
In a PSR, we similarly allow any feature function that is linear in the predictive state vector $q\in \mathbb R^k$: 
$$f(q,a) = F_a q$$
with one matrix $F_a\in\mathbb R^{d\times k}$ for each action $a$. In either case, define the successor features to be
$$\phi^\pi(q) = \mathbb E_\pi\left[\sum_{t=1}^\infty \gamma^{t-1} f(q_t, a_t) \,\bigg|\, \mathrm{do}~q_1=q \right]$$
Interestingly, the function $\phi^\pi$ is \emph{linear} in $q$. That is, for each $\pi$, there exists a matrix $A^\pi\in\mathbb R^{d\times k}$ such that $\phi^\pi(q) = A^\pi q$. We call $A^\pi$ the \emph{successor feature matrix} for $\pi$; it is related to the parameters of the \emph{Linear Successor Feature Model} of \citet{lehnert19lsfm}.

We can compute $A^\pi$ recursively by working backward in time (upward from the leaves of a policy tree): for a tree with root action $a$, the recursion is
$$A^\pi = F_a + \gamma \sum_o A^{\pi(o)} T_{ao}$$
This recursion works by splitting $A^\pi$ into contributions from the first step ($F_a$) and from steps $2\ldots H$ (rest of RHS). We give a more detailed derivation, as well as a proof of linearity, in the supplementary material online. %
All the above works for MDPs as well by taking $q_t = e_{s_t}$, which lets us keep a uniform notation across MDPs, POMDPs, and PSRs.

It is worth noting the multiple feature representations that contribute to the function $\phi^\pi(q)$. First are the immediate features $f(q,a)$. Second is the PSR state, which can often be thought of as a feature representation for an underlying ``uncompressed'' model \citep{hefny15psr}. Finally, both of the above feature representations help define the \emph{exact} value of $\phi^\pi$; we can also \emph{approximate} $\phi^\pi$ using a third feature representation. Any of these feature representations could be related, or we could use separate features for all three purposes. We believe that an exploration of the roles of these different representations would be important and interesting, but we leave it for future work.

\section{Successor Feature Sets}

To reason about multiple policies, we can collect together multiple matrices: the \emph{successor feature set} at horizon $H$ is defined as the set of all possible successor feature matrices at horizon $H$,
$$\Phi^{(H)} = \{ A^\pi \mid \pi~\text{a policy with horizon}~H \}$$
As we will detail below, we can also define an infinite-horizon successor feature set $\Phi$, which is the limit of $\Phi^{(H)}$ as $H\to\infty$. 

The successor feature set tells us \emph{how the future depends} on our state and our choice of policy. It tells us the range of outcomes that are possible: for a state $q$, each point in $\Phi q$ tells us about one policy, and gives us moments of the distribution of future states under that policy. The extreme points of $\Phi q$ therefore tell us the limits of what we can achieve. (Here we use the shorthand of \emph{broadcasting}: set arguments mean that we perform an operation all possible ways, substituting one element from each set. E.g., if $X,Y$ are sets, $X+Y$ means Minkowski sum $\{x+y \mid x\in X, y\in Y\}$.) 

Note that $\Phi^{(H)}$ is a convex, compact set: by linearity of expectation, the feature matrix for a stochastic policy will be a convex combination of the matrices for its component deterministic policies. Therefore, $\Phi^{(H)}$ will be the convex hull of a finite set of matrices, one for each possible deterministic policy at horizon $H$.

Working with multiple policies at once provides a number of benefits: perhaps most importantly, it lets us define a Bellman backup that builds new policies combinatorially by combining existing policies at each iteration (Sec.~\ref{sec:all-policy-bellman}). That way, we can reason about all possible policies instead of just a fixed list. Another benefit of $\Phi$ is that, as we will see below, it can help us compute optimal policies and feature-matching policies efficiently.
On the other hand, because it contains so much information, the set $\Phi$ is a complicated object; it can easily become impractical to work with. We return to this problem in Sec.~\ref{sec:practical-implementation}.

\section{Special Cases}

\begin{figure}[t]
\centering
\includegraphics[width=0.7\linewidth]{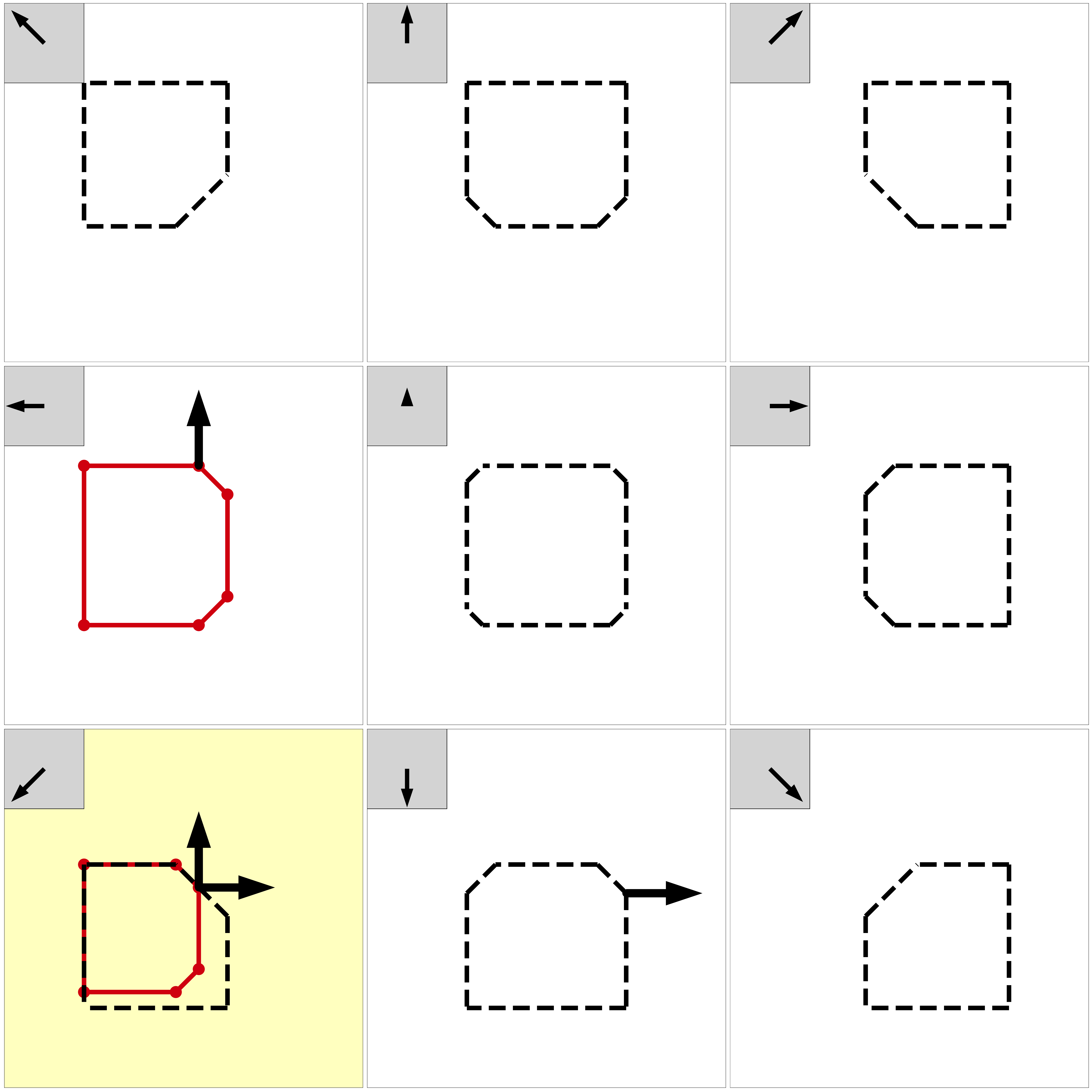}
\label{fig:gridworld}
\caption{Visualization of the successor feature set $\Phi$ for a $3\times 3$ gridworld MDP with 2d features. Start state is in yellow. Gray insets show one-step feature vectors, which depend only on the state, not the action. Each subplot shows one projection $\Phi e_j$ (scale is arbitrary, so no axes are necessary). The red sets illustrate a Bellman backup at the bottom-left state, and the black arrows illustrate the feature-matching policy there. See text for details.}
\label{fig:gridworld_fig}
\end{figure}

In some useful special cases, successor feature matrices and successor feature sets have a simpler structure that can make them easier to reason about and work with.  
E.g., in an MDP, we can split the successor feature matrix into its columns, resulting in one vector per state --- this is the ordinary successor feature vector $\phi^\pi(s) = A^\pi e_s$.
Similarly, we can split $\Phi$ into sets of successor feature vectors, one at each state, representing the range of achievable futures:
$$\phi(s) = \{ \phi^\pi(s) \mid \pi~\text{a policy}\} = \Phi e_s$$
Fig.~\ref{fig:gridworld_fig} visualizes these projections, along with the Bellman backups described below. Each projection tells us the discounted total feature vectors that are achievable from the corresponding state. For example, the top-left plot shows a set with five corners, each corresponding to a policy that is optimal in this state under a different reward function; the bottom-left corner corresponds to ``always go down,'' which is optimal under reward $R(s,a) = (-1, -1) f(s,a)$.

On the other hand, if we only have a single one-step feature ($f(q,a)\in\mathbb R)$, then we can only represent a 1d family of reward functions. All positive multiples of $f$ are equivalent to one another, as are all negative multiples. In this case, our recursion effectively reduces to classic POMDP or PSR value iteration: each element of $\Phi$ is now a vector $\alpha^\pi\in\mathbb R^k$ instead of a matrix $A^\pi\in\mathbb R^{d\times k}$. This $\alpha$-vector represents the (linear) value function of policy $\pi$; the pointwise maximum of all these functions is the (piecewise linear and convex) optimal value function of the POMDP or PSR.

\section{Bellman Equations}
\label{sec:all-policy-bellman}

Each element of the successor feature set is a successor feature matrix for some policy, and as such, it satisfies the recursion given above. For efficiency, though, we would like to avoid running Bellman backups separately for too many possible policies. To this end, we can write a backup operator and Bellman equations that apply to all policies at once, and hence describe the entire successor feature set. 

The joint backup works by relating horizon-$H$ policies to horizon-$(H-1)$ policies. Every horizon-$H$ policy tree can be constructed recursively, by choosing an action to perform at the root node and a horizon-$(H-1)$ tree to execute after each possible observation. So, we can break down any horizon-$H$ policy (including stochastic ones) into a distribution over the initial action, followed by conditional distributions over horizon-$(H-1)$ policy trees for each possible initial observation.

Therefore, if we have the successor feature set $\Phi^{(H-1)}$ at horizon $H-1$, we can construct the successor feature set at horizon $H$ in two steps: first, for each possible initial action $a$, we construct
$$\Phi^{(H)}_a = F_a + \gamma \sum_o \Phi^{(H-1)} T_{ao}$$
This set tells us the successor feature matrices for all horizon-$H$ policies that begin with action $a$. Note that only the first action is deterministic: $\Phi^{(H-1)}$ lets us assign any conditional distribution over horizon-$(H-1)$ policy trees after each possible observation.

Second, since a general horizon-$H$ policy is a distribution over horizon-$H$ policies that start with different actions, each element of $\Phi^{(H)}$ is a convex combination of elements of $\Phi^{(H)}_a$ for different values of $a$. That is,
$$\Phi^{(H)} = \mathrm{conv} \bigcup_a \Phi^{(H)}_a $$
The recursion bottoms out at horizon $0$, where we have
$$\Phi^{(0)} = \{0\}$$
since the discounted sum of a length-$0$ trajectory is always the zero vector.

Fig.~\ref{fig:gridworld_fig} shows a simple example of the Bellman backup. Since this is an MDP, $\Phi$ is determined by its projections $\Phi e_j$ onto the individual states. The action ``up'' takes us from the bottom-left state to the middle-left state. So, we construct $\Phi_{\text{up}} e_{\text{bottom-left}}$ by shifting and scaling $\Phi e_{\text{middle-left}}$ (red sets). The full set $\Phi e_{\text{bottom-left}}$ is the convex hull of four sets $\Phi_{a} e_{\text{bottom-left}}$; the other three are not shown, but for example, taking $a=\mathrm{right}$ gives us a shifted and scaled copy of the set from the bottom-center plot.

The update from $\Phi^{(H-1)}$ to $\Phi^{(H)}$ is a contraction: see %
the supplementary material online for a proof.
So, as $H\to\infty$, $\Phi^{(H)}$ will approach a limit $\Phi$; this set represents the achievable successor feature matrices in the infinite-horizon discounted setting. $\Phi$ is a fixed point of the Bellman backup, and therefore satisfies the \emph{stationary} Bellman equations
$$
\Phi = \mathrm{conv} \bigcup_a \left[ F_a + \gamma \sum_o \Phi T_{ao} \right]
$$

\section{Feature Matching and Optimal Planning}

Once we have computed the successor feature set, we can return to the feature matching task described in Section~\ref{sec:feature-matching}. Knowing $\Phi$ makes feature matching easier: for any target vector of discounted feature expectations $\phi^d$, we can efficiently either compute a policy that matches $\phi^d$ or verify that matching $\phi^d$ is impossible. We detail an algorithm for doing so in Alg.~\ref{alg-featmat}; more detail is in the supplementary material.

Fig.~\ref{fig:gridworld_fig} shows the first steps of our feature-matching policy in a simple MDP\@. At the bottom-left state, the two arrows show the initial target feature vector (root of the arrows) and the computed policy (randomize between ``up'' and ``right'' according to the size of the arrows). The target feature vector at the next step depends on the outcome of randomization: each destination state shows the corresponding target and the second step of the computed policy.

\begin{algorithm}
\caption{Feature Matching Policy}\label{alg-featmat}
\DontPrintSemicolon
\SetAlgoLined
$t \leftarrow 1$\;
Initialize $\phi^d_t$ to the target vector of expected discounted features.\;
Initialize $q_t$ to the initial state of the environment.\;
\Repeat{done}{
  Choose actions $a_{it}$, vectors $\phi_{it}\in\Phi_{a_{it}} q_t$, and convex combination weights $p_{it}$ s.t. $\phi^d_t = \sum_{i=1}^\ell p_{it} \phi_{it}$.\;
  Choose an index $i$ according to probabilities $p_{it}$, and execute the corresponding action: $a_t \leftarrow a_{it}$.\;
  Write the corresponding $\phi_{it}$ as 
$\phi_{it} = F_{a_{t}} q_t + \gamma\sum_o \phi_{ot}$ by choosing $\phi_{ot}\in \Phi T_{a_{t}o}q_t$ for each $o$.\;
  Receive observation $o_t$, and calculate $p_{t} = P(o_t\mid q_t, a_t) = u^TT_{a_to_t}q_t$.\;
  $q_{t+1} \leftarrow T_{a_{t}o_{t}} q_t / p_{t}$\;
  $\phi^d_{t+1} \leftarrow \phi_{o_tt}/p_{t}$\;
  $t \leftarrow t + 1$\;
}
\end{algorithm}

We can also use the successor feature set to make optimal planning easier. In particular, if we are given a new reward function expressed in terms of our features, say $R(q,a) = r^T f(q,a)$ for some coefficient vector $r$, then we can efficiently compute the optimal value function under $R$:
$$V^*(q) = \max_\pi r^T \phi^\pi q = \max\, \{ r^T \psi q \mid \psi\in \Phi \}$$
As a by-product we get an optimal policy: there will always be a matrix $\psi$ that achieves the $\max$ above and satisfies $\psi\in\Phi_a$ for some $a$. Any such $a$ is an optimal action.

\section{Implementation}
\label{sec:practical-implementation}

An exact representation of $\Phi$ can grow faster than exponentially with the horizon. So, in our experiments below, we work with a straightforward approximate representation. We use two tools: 
first, we store $\Phi_{ao} = \Phi T_{ao}$ for all $a,o$ instead of storing $\Phi$, since the former sets tend to be effectively lower-dimensional due to sparsity. Second, analogous to PBVI \citep{Pineau03apbvi, shani13pbvi}, we fix a set of directions $m_i\in\mathbb R^{d\times k}$, and retain only the most extreme point of $\Phi_{ao}$
in each direction. Our approximate backed-up set is then the convex hull of these retained points. Just as in PBVI, we can efficiently compute backups by passing the max through the Minkowski sum in the Bellman equation. That is, for each $i$ and each $a,o$, we solve
$$\textstyle \arg\max\, \langle m_i , \phi\rangle \mbox{ for } \phi \in \bigcup_{a'} \left[ F_{a'} + \gamma\sum_{o'} \Phi_{a'o'}\right] T_{ao} $$
by solving, for each $i,a,o,a',o'$
$$\textstyle \arg\max\, \langle m_i , \phi \rangle \mbox{ for } \phi \in \Phi_{a'o'}T_{ao} $$
and combining the solutions.

There are a couple of useful variants of this implementation that we can use in \emph{stoppable} problems (i.e., problems where we have an emergency-stop or a safety policy; see the supplemental material for more detail). First, we can update \emph{monotonically}, i.e., keep the better of the horizon-$H$ or horizon-$(H+1)$ successor feature matrices in each direction. Second, we can update \emph{incrementally}: we can update any subset of our directions while leaving the others fixed.

\begin{figure*}[ht]
    \centering
    \begin{subfigure}[t]{1.0\textwidth}
    \centerline{\includegraphics[width=1.0\textwidth]{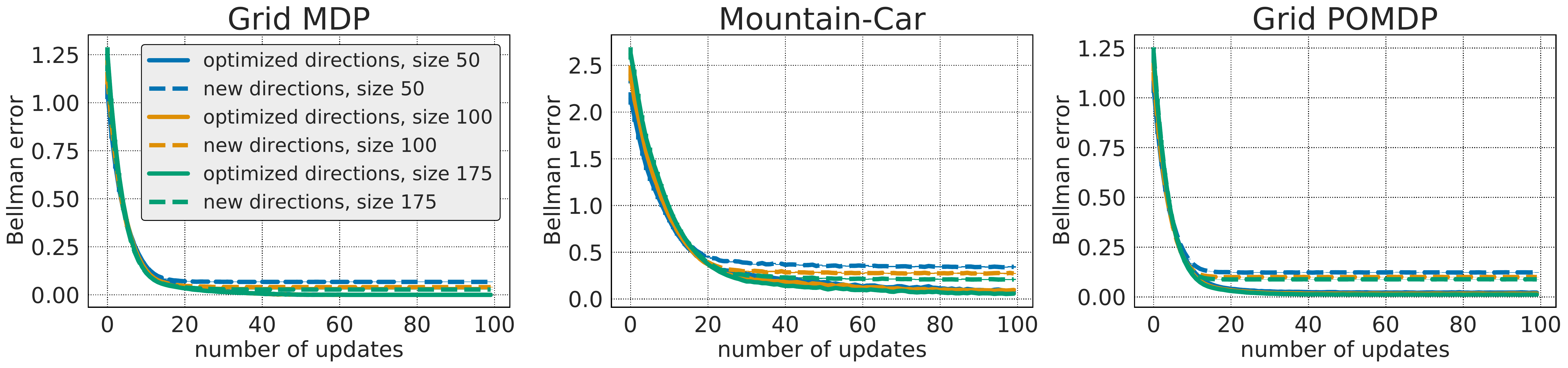}}
    \end{subfigure}
    \caption{%
    Bellman error v.\ iteration for three simple test domains, varying the amount of computation per iteration. We show error separately in directions we have optimized over and in new random directions. Average of 25 random seeds of the {direction $m_{i}$ with the highest bellman error per seed}; all error bars %
    are smaller than the line widths. The center panel shows the effect on Bellman error when we have higher-dimensional feature vectors. The rightmost panel shows the effect on Bellman error when the agent has less information about the exact state. In both cases the convergence rate stays similar, but we need more directions $m_i$ to adequately sample the boundary of $\Phi$ (i.e., to lower the asymptotic error on new directions).}
    \label{fig:convergence_fig}
\end{figure*}

\section{More on Special Cases}

With the above pruning strategy, our dynamic programming iteration generalizes PBVI~\cite{pineau03pbvi}.
PBVI was defined originally for POMDPs, but it extends readily to PSRs as well: we just sample predictive states instead of belief states. To relate PBVI to our method, we look at a single task, with reward coefficient vector $r$. We sample a set of belief states or predictive states $q_i$; these are the directions that PBVI will use to decide which value functions ($\alpha$-vectors) to retain. Based on these, we set the successor feature matrix directions to be $m_i = r q_i^T$ for all $i$. 

Now, when we search within our backed up set $\Phi^{(H)}$ for the maximal element in direction $m_i$, we get some successor feature matrix $\phi$. Because $\text{tr}(\phi^T r q_i^T)$ is maximal, we know that $\text{tr}(q_i^T \phi^T r) = q_i^T (\phi^T r)$ is also maximal: that is, $\phi^T r$ is as far as possible in the direction $q_i$. But $\phi^T r$ is a backed-up value function under the reward $r$; so, $\phi^T r$ is exactly the value function that PBVI would retain when maximizing in the direction $q_i$.

\section{Experiments: Dynamic Programming}

We tried our dynamic programming method on several small domains: the classic mountain-car domain and a random $18\times 18$ gridworld with full and partial observability. We evaluated both planning and feature matching; results for the former are discussed in this section, and an example of the latter is in Fig.~\ref{fig:gridworld_fig}.
We give further details of our experimental setup in the supplementary material online. %
At a high level, our experiments show that the algorithms behave as expected, and that they are practical for small domains. They also tell us about limits on scaling: the tightest of these limits is our ability to represent $\Phi$ accurately, governed by the number of boundary points that we retain for each $\Phi_{ao}$.

In mountain-car, the agent has two actions: accelerate left and accelerate right. The state is (position, velocity), in $[-1.2, 0.6]\times [ -0.07, 0.07 ]$. We discretize to a $12\times 12$ mesh with piecewise-constant approximation. 
Our one-step features are radial basis functions of the state, with values in $[0,1]$. 
We use 9 RBF centers evenly spaced on a $3\times 3$ grid.
In the MDP gridworld, the agent has four deterministic actions: up, down, left, and right. The one-step features are $(x,y)$ coordinates scaled to $[-1, 1]$, similar to Fig.~\ref{fig:gridworld_fig}. In the POMDP gridworld, the actions are stochastic, and the agent only sees a noisy indicator of state. In all domains, the discount is $\gamma=0.9$.

Fig.~\ref{fig:convergence_fig} shows how Bellman error evolves across iterations of dynamic programming. Since $\Phi$ is a set, we evaluate error by looking at random projections: how far do $\Phi$ and the backup of $\Phi$ extend in a given direction? We evaluate directions $m_i$ that we optimized for during backups, as well as new random directions. 

Note that the asymptotes for the new-directions lines are above zero; this persistent error is due to our limited-size representation of $\Phi$. The error decreases as we increase the number of boundary points that we store. It is larger in the domains with more features and more uncertainty (center and right panels), due to the higher-dimensional $A^\pi$ matrices and the need to sample mixed (uncertain) belief states.

\section{Related Work}
Successor features, a version of which were first introduced by \citet{dayan1993improving}, provide a middle ground between model-free and model-based RL \citep{russek2017predictive}.
They have been proposed as neurally plausible explanations of learning \citep{gershman2012successor,gershman2018successor,momennejad2017successor,stachenfeld2017hippocampus,gardner2018rethinking,vertes2019neurally}.

Recently, numerous extensions have been proposed.
Most similar to the current work are methods that generalize to a set of policies or tasks. \citet{barreto2017successor} achieve transfer learning by generalizing across tasks with successor features; \citet{barreto2018transfer} use generalized policy improvement (GPI)  over a set of policies. A few methods \citep{borsa2018universal,ma2018universal} recently combined universal value function approximators \citep{schaul2015universal} with GPI to perform multi-task learning, generalizing to a set of goals by conditioning on a goal representation. \citet{barreto2020fast} extend policy improvement and policy evaluation from single tasks and policies to a list of them, but do not attempt to back up across policies.

Many authors have trained nonlinear models such as neural networks to predict successor-style representations, e.g., \citet{kulkarni2016deep,zhu2017visual,zhang2017deep,machado2017eigenoption,hansen2019fast}. These works are complementary to our goal here, which is to design and analyze new, more general successor-style representations. We hope our generalizations eventually inform training methods for large-scale nonlinear models.

At the intersection of successor features and imitation learning, \citet{zhu2017visual} address visual semantic planning; \citet{lee2019truly} address off-policy model-free RL in a batch setting; and \citet{hsu2019new} addresses active imitation learning.

As mentioned above, the individual elements of $\Phi$ are related to the work of \citet{lehnert19lsfm}. And, we rely on point-based methods \citep{Pineau03apbvi, shani13pbvi} to compute $\Phi$.

\section{Conclusion}
This work introduces \emph{successor feature sets}, a new representation that generalizes successor features. Successor feature sets represent and reason about successor feature predictions for all policies at once, and respect the compositional structure of policies, in contrast to other approaches that treat each policy separately. The set represents the boundaries of what is achievable in the future, and how these boundaries depend on our initial state. This information lets us read off optimal policies or imitate a demonstrated behavior.

We give algorithms for working with successor feature sets, including a dynamic programming algorithm to compute them, as well as algorithms to read off policies from them. The dynamic programming update is a contraction mapping, and therefore convergent. We give both exact and approximate versions of the update. The exact version can be intractable, due to the so-called ``curse of dimensionality'' and ``curse of history.'' The approximate version mitigates these curses using point-based sampling. 

Finally, we present computational experiments. These are limited to relatively small, known environments; but in these environments, we demonstrate that we can compute successor feature sets accurately, and that they aid generalization.
We also explore how our approximations scale with environment complexity.

Overall we believe that our new representation can provide insight on how to reason about policies in a dynamical system.
We know, though, that we have only scratched the surface of possible strategies for working with this representation, and we hope that our analysis can inform future work on larger-scale environments.

\bibliography{bibliography}

\begin{thebibliography}{28}
\providecommand{\natexlab}[1]{#1}
\providecommand{\url}[1]{\texttt{#1}}
\providecommand{\urlprefix}{URL }
\expandafter\ifx\csname urlstyle\endcsname\relax
  \providecommand{\doi}[1]{doi:\discretionary{}{}{}#1}\else
  \providecommand{\doi}{doi:\discretionary{}{}{}\begingroup
  \urlstyle{rm}\Url}\fi

\bibitem[{Abbeel and Ng(2004)}]{abbeel2004apprenticeship}
Abbeel, P.; and Ng, A.~Y. 2004.
\newblock Apprenticeship learning via inverse reinforcement learning.
\newblock In \emph{Proceedings of the twenty-first international conference on
  Machine learning}, 1.

\bibitem[{Barreto et~al.(2018)Barreto, Borsa, Quan, Schaul, Silver, Hessel,
  Mankowitz, Zidek, and Munos}]{barreto2018transfer}
Barreto, A.; Borsa, D.; Quan, J.; Schaul, T.; Silver, D.; Hessel, M.;
  Mankowitz, D.; Zidek, A.; and Munos, R. 2018.
\newblock Transfer in deep reinforcement learning using successor features and
  generalised policy improvement.
\newblock In \emph{International Conference on Machine Learning}, 501--510.
  PMLR.

\bibitem[{Barreto et~al.(2017)Barreto, Dabney, Munos, Hunt, Schaul, van
  Hasselt, and Silver}]{barreto2017successor}
Barreto, A.; Dabney, W.; Munos, R.; Hunt, J.~J.; Schaul, T.; van Hasselt,
  H.~P.; and Silver, D. 2017.
\newblock Successor features for transfer in reinforcement learning.
\newblock In \emph{Advances in neural information processing systems},
  4055--4065.

\bibitem[{Barreto et~al.(2020)Barreto, Hou, Borsa, Silver, and
  Precup}]{barreto2020fast}
Barreto, A.; Hou, S.; Borsa, D.; Silver, D.; and Precup, D. 2020.
\newblock Fast reinforcement learning with generalized policy updates.
\newblock \emph{Proceedings of the National Academy of Sciences} .

\bibitem[{Borsa et~al.(2018)Borsa, Barreto, Quan, Mankowitz, Munos, van
  Hasselt, Silver, and Schaul}]{borsa2018universal}
Borsa, D.; Barreto, A.; Quan, J.; Mankowitz, D.; Munos, R.; van Hasselt, H.;
  Silver, D.; and Schaul, T. 2018.
\newblock Universal successor features approximators.
\newblock \emph{arXiv preprint arXiv:1812.07626} .

\bibitem[{Dayan(1993)}]{dayan1993improving}
Dayan, P. 1993.
\newblock Improving generalization for temporal difference learning: The
  successor representation.
\newblock \emph{Neural Computation} 5(4): 613--624.

\bibitem[{Gardner, Schoenbaum, and Gershman(2018)}]{gardner2018rethinking}
Gardner, M.~P.; Schoenbaum, G.; and Gershman, S.~J. 2018.
\newblock Rethinking dopamine as generalized prediction error.
\newblock \emph{Proceedings of the Royal Society B} 285(1891): 20181645.

\bibitem[{Gershman(2018)}]{gershman2018successor}
Gershman, S.~J. 2018.
\newblock The successor representation: its computational logic and neural
  substrates.
\newblock \emph{Journal of Neuroscience} 38(33): 7193--7200.

\bibitem[{Gershman et~al.(2012)Gershman, Moore, Todd, Norman, and
  Sederberg}]{gershman2012successor}
Gershman, S.~J.; Moore, C.~D.; Todd, M.~T.; Norman, K.~A.; and Sederberg, P.~B.
  2012.
\newblock The successor representation and temporal context.
\newblock \emph{Neural Computation} 24(6): 1553--1568.

\bibitem[{Hansen et~al.(2019)Hansen, Dabney, Barreto, Van~de Wiele,
  Warde-Farley, and Mnih}]{hansen2019fast}
Hansen, S.; Dabney, W.; Barreto, A.; Van~de Wiele, T.; Warde-Farley, D.; and
  Mnih, V. 2019.
\newblock Fast task inference with variational intrinsic successor features.
\newblock \emph{arXiv preprint arXiv:1906.05030} .

\bibitem[{Hefny, Downey, and Gordon(2015)}]{hefny15psr}
Hefny, A.; Downey, C.; and Gordon, G. 2015.
\newblock Supervised Learning for Dynamical System Learning.
\newblock In \emph{Advances in neural information processing systems}.

\bibitem[{Hsu(2019)}]{hsu2019new}
Hsu, D. 2019.
\newblock A New Framework for Query Efficient Active Imitation Learning.
\newblock \emph{arXiv preprint arXiv:1912.13037} .

\bibitem[{Jaeger(2000)}]{jaeger00ooms}
Jaeger, H. 2000.
\newblock Observable operator models for discrete stochastic time series.
\newblock \emph{Neural Computation} 12(6): 1371--1398.

\bibitem[{Kulkarni et~al.(2016)Kulkarni, Saeedi, Gautam, and
  Gershman}]{kulkarni2016deep}
Kulkarni, T.~D.; Saeedi, A.; Gautam, S.; and Gershman, S.~J. 2016.
\newblock Deep successor reinforcement learning.
\newblock \emph{arXiv preprint arXiv:1606.02396} .

\bibitem[{Lee, Srinivasan, and Doshi-Velez(2019)}]{lee2019truly}
Lee, D.; Srinivasan, S.; and Doshi-Velez, F. 2019.
\newblock Truly batch apprenticeship learning with deep successor features.
\newblock \emph{arXiv preprint arXiv:1903.10077} .

\bibitem[{Lehnert and Littman(2019)}]{lehnert19lsfm}
Lehnert, L.; and Littman, M.~L. 2019.
\newblock Successor Features Combine Elements of Model-Free and Model-based
  Reinforcement Learning.
\newblock Technical Report arXiv:1901.11437.

\bibitem[{Ma, Wen, and Bengio(2018)}]{ma2018universal}
Ma, C.; Wen, J.; and Bengio, Y. 2018.
\newblock Universal successor representations for transfer reinforcement
  learning.
\newblock \emph{arXiv preprint arXiv:1804.03758} .

\bibitem[{Machado et~al.(2017)Machado, Rosenbaum, Guo, Liu, Tesauro, and
  Campbell}]{machado2017eigenoption}
Machado, M.~C.; Rosenbaum, C.; Guo, X.; Liu, M.; Tesauro, G.; and Campbell, M.
  2017.
\newblock Eigenoption discovery through the deep successor representation.
\newblock \emph{arXiv preprint arXiv:1710.11089} .

\bibitem[{Momennejad et~al.(2017)Momennejad, Russek, Cheong, Botvinick, Daw,
  and Gershman}]{momennejad2017successor}
Momennejad, I.; Russek, E.~M.; Cheong, J.~H.; Botvinick, M.~M.; Daw, N.~D.; and
  Gershman, S.~J. 2017.
\newblock The successor representation in human reinforcement learning.
\newblock \emph{Nature Human Behaviour} 1(9): 680--692.

\bibitem[{Pineau, Gordon, and Thrun(2003{\natexlab{a}})}]{Pineau03apbvi}
Pineau, J.; Gordon, G.; and Thrun, S. 2003{\natexlab{a}}.
\newblock Point-based Value Iteration: An Anytime Algorithm for {POMDPs}.
\newblock In \emph{Proceedings of the Sixteenth International Joint Conference
  on Artificial Intelligence (IJCAI)}.

\bibitem[{Pineau, Gordon, and Thrun(2003{\natexlab{b}})}]{pineau03pbvi}
Pineau, J.; Gordon, G.~J.; and Thrun, S.~B. 2003{\natexlab{b}}.
\newblock In \emph{Proceedings of the 18th International Joint Conference on
  Artificial Intelligence (IJCAI)}, 1025--1030.

\bibitem[{Russek et~al.(2017)Russek, Momennejad, Botvinick, Gershman, and
  Daw}]{russek2017predictive}
Russek, E.~M.; Momennejad, I.; Botvinick, M.~M.; Gershman, S.~J.; and Daw,
  N.~D. 2017.
\newblock Predictive representations can link model-based reinforcement
  learning to model-free mechanisms.
\newblock \emph{PLoS computational biology} 13(9): e1005768.

\bibitem[{Schaul et~al.(2015)Schaul, Horgan, Gregor, and
  Silver}]{schaul2015universal}
Schaul, T.; Horgan, D.; Gregor, K.; and Silver, D. 2015.
\newblock Universal value function approximators.
\newblock In \emph{International conference on machine learning}, 1312--1320.

\bibitem[{Shani, Pineau, and Kaplow(2013)}]{shani13pbvi}
Shani, G.; Pineau, J.; and Kaplow, R. 2013.
\newblock A survey of point-based POMDP solvers.
\newblock \emph{Autonomous Agents and Multi-Agent Systems} 27: 1--51.

\bibitem[{Stachenfeld, Botvinick, and
  Gershman(2017)}]{stachenfeld2017hippocampus}
Stachenfeld, K.~L.; Botvinick, M.~M.; and Gershman, S.~J. 2017.
\newblock The hippocampus as a predictive map.
\newblock \emph{Nature neuroscience} 20(11): 1643.

\bibitem[{V{\'e}rtes and Sahani(2019)}]{vertes2019neurally}
V{\'e}rtes, E.; and Sahani, M. 2019.
\newblock A neurally plausible model learns successor representations in
  partially observable environments.
\newblock In \emph{Advances in Neural Information Processing Systems},
  13714--13724.

\bibitem[{Zhang et~al.(2017)Zhang, Springenberg, Boedecker, and
  Burgard}]{zhang2017deep}
Zhang, J.; Springenberg, J.~T.; Boedecker, J.; and Burgard, W. 2017.
\newblock Deep reinforcement learning with successor features for navigation
  across similar environments.
\newblock In \emph{2017 IEEE/RSJ International Conference on Intelligent Robots
  and Systems (IROS)}, 2371--2378. IEEE.

\bibitem[{Zhu et~al.(2017)Zhu, Gordon, Kolve, Fox, Fei-Fei, Gupta, Mottaghi,
  and Farhadi}]{zhu2017visual}
Zhu, Y.; Gordon, D.; Kolve, E.; Fox, D.; Fei-Fei, L.; Gupta, A.; Mottaghi, R.;
  and Farhadi, A. 2017.
\newblock Visual semantic planning using deep successor representations.
\newblock In \emph{Proceedings of the IEEE international conference on computer
  vision}, 483--492.

\end{thebibliography}

\clearpage
\appendix

\section*{Supplementary material}

\section{Feature matching}

In this section we give the algorithm for imitation by feature matching, summarized as Alg.~\ref{alg-featmat}.

Our policy will be nonstationary: that is, its actions will depend on an internal policy state (defined below) as well as the environment's current predictive state $q_t$.

Our algorithm updates its target feature vector over time in order to compensate for random outcomes (the action sampled from the policy and the next state sampled from the transition distribution). We write $\phi^d_t$ for the target at time step $t$, and initialize $\phi^d_1=\phi^d$. Updates of this sort are necessary: we might by chance visit a state where it is impossible to achieve the original target $\phi^d_1$, but that does not mean that our policy has failed. Instead, the policy guarantees always to pick a target $\phi^d_t$ that is achievable given the state $q_t$ at step $t$, in a way that guarantees that on average we achieve the original target $\phi^d_1$ from the initial state $q_1$.

To guarantee that the target is always achievable, our policy maintains the invariant that $\phi^d_t\in\Phi q_t$. By the definition of $\Phi$, the discounted feature vectors in $\Phi q$ are exactly the ones that are achievable starting from state $q$, so this invariant is necessary and sufficient to ensure that $\phi^d_t$ is achievable. At the first time step, we test whether $\phi^d_1\in\Phi q_1$. If yes, our invariant is satisfied and we can proceed; if no, then we know that we have been given an impossible task. In the latter case we could raise an error, or we could raise a warning and look for the closest achievable vector $\phi\in\Phi q$ to $\phi^d_1$.

Our actions at step $t$ and our targets at step $t+1$ will be functions of the current environment state $q_t$ and our current target $\phi^d_t$. As such, $\phi^d_t$ is the internal policy state mentioned above.

We pick our actions and targets as follows.
According to the Bellman equations, the successor feature set $\Phi$ is equal to the convex hull of the union of $\Phi_a$ over all $a$. Each matrix in $\Phi$ can therefore be written as a convex combination of action-specific matrices, each one chosen from one of the sets $\Phi_a$. That means that each vector in $\Phi q_t$ can be written as a convex combination of vectors in $\Phi_a q_t$.

Write our target $\phi^d_t$ in this way, say $\phi^d_t = \sum_{i=1}^\ell p_{it} \phi_{it}$, by choosing actions $a_{it}$, vectors $\phi_{it}\in\Phi_{a_{it}} q_t$, and weights $p_{it}\geq 0$ with $\sum_i p_{it}=1$. Then, at the current time step, our algorithm chooses an index $i$ according to the probabilities $p_{it}$, and executes the corresponding action $a_{it}$.

Now let $i$ be the chosen index, and write $a=a_{it}$ for the chosen action. Again according to the Bellman equations, the point $\phi_{it}$ is of the form $[F_{a} + \gamma \sum_{o} \Phi T_{ao}] q_t$. In particular, we can choose vectors $\phi_{ot}\in \Phi T_{ao}q_t$ for each $o$ such that
$$\phi_{it} = F_a q_t + \gamma\sum_o \phi_{ot}$$
Writing $p_{ot} = P(o\mid q_t, a)$ for all $o$, we can multiply and divide by $p_{ot}$ within the sum, and conclude
$$\phi_{it} = \mathbb E_o[F_a q_t + \gamma \phi_{ot}/p_{ot} ]$$
That is, we can select our target for time step $t+1$ as $\phi^d_{t+1} = \phi_{ot}/p_{ot}$, where $o=o_t$ is our next observation. To see why, note that our expected discounted feature vector at time $t$ will remain the same: the LHS (the current target) is equal to the RHS (the expected one-step contribution plus discounted future target). And, note that the target at the next time step will always be feasible, maintaining our invariant: our state at the next time step will be
$$q_{t+1} = T_{ao} q_t / p_{ot}$$
and we have selected each $\phi_{ot}$ to satisfy $\phi_{ot}\in \Phi T_{ao}q_t$, so
$$\phi_{ot} / p_{ot} \in \Phi T_{ao} q_t / p_{ot} = \Phi q_{t+1}$$
So, based on the observation that we receive, we can update our predictive state and target feature vector according to the equations above, and recurse. (In practice, numerical errors or incomplete convergence of $\Phi$ could lead to an infeasible target; in this case we can project the target back onto the feasible set, which will result in some error in feature matching.)

Note that there may be more than one way to decompose $\phi^d_t = \sum_{i=1}^\ell p_{it} \phi_{it}$, or more than one way to decompose $\phi_{it} = F_a q_t + \gamma\sum_o \phi_{ot}$. If so, we can choose any valid decomposition arbitrarily.

\section{Convergence of dynamic programming}
\label{sec:convergence}

We will show that the dynamic programming update for $\Phi$ given in Section~\ref{sec:all-policy-bellman} is a contraction, which implies bounds on the convergence rate of dynamic programming.
We will need a few definitions and facts about norms and metrics.

\subsection{Norms}

Given any symmetric, convex, compact set $S$ with nonempty interior, we can construct a norm by treating $S$ as the unit ball. The norm of a vector $x$ is then the smallest multiple of $S$ that contains $x$.
$$\|x\|_S = \inf_{x\in cS} c$$
This is a fully general way to define a norm: any norm can be constructed this way by using its own unit ball as $S$. That is, if $B = \{x \mid \|x\|\leq 1\}$, then
$$\|x\| = \|x\|_B$$
We will use the shorthand $\|\cdot\|_p$ for an $L_p$-norm: e.g., $L_1$, $L_2$ (Euclidean norm), or $L_\infty$ (sup norm).
If we start from an asymmetric set $S$, we can symmetrize it to get
$$\bar S = \{ \alpha s + (1-\alpha) s' \mid s\in S, s'\in -S, \alpha\in[0,1] \}$$
(This is the convex hull of $S\cup -S$.)
Given any norm $\|\cdot\|$, we can construct a \emph{dual norm} $\|\cdot\|^*$:
$$\|y\|^* = \sup_{\|x\|\leq 1} x\cdot y$$
This definition guarantees that dual norms satisfy H\"older's inequality:
$$x\cdot y \leq \|x\|\,\|y\|^*$$
We will write $S^*$ for the unit ball of the dual norm $\|\cdot\|^*_{S}$. Taking the dual twice returns to the original norm: $\|\cdot\|_S^{**}=\|\cdot\|_S$ and $S^{**}=S$.

Given any two norms $\|\cdot\|_P$ and $\|\cdot\|_Q$ and their corresponding unit balls $P$ and $Q$, the \emph{operator norm} of a matrix $A$ is
$$\|A\|_{P,Q} = \sup_{x\in P} \|Ax\|_Q = \sup_{x\in P, y\in Q^*} y^TAx$$
This definition ensures that H\"older's inequality extends to operator norms:
$$\|Ax\|_{Q} \leq \|A\|_{P,Q} \|x\|_P$$
The norm of the transpose of a matrix can be expressed in terms of the duals of $\|\cdot\|_P$ and $\|\cdot\|_Q$:
$$\|A^T\|_{Q^*,P^*} = \|A\|_{P,Q}$$
If $P$ and $Q$ are the same, we will shorten to 
$$\|A^T\|_{P^*} = \|A\|_P$$
Given a norm, we can define the Hausdorff metric between sets:
\begin{align*}
    d(X, Y) & = \max(\bar d(X, Y), \bar d(Y, X)) \\
    \bar d(X, Y) &= \sup_{x\in X}\inf_{y\in Y} \|x-y\|
\end{align*}
If $V$ is any real vector space (such as $\mathbb R^{d\times k}$), the Hausdorff metric makes the set of non-empty compact subsets of $V$ into a complete metric space.
Given a metric, a \emph{contraction} is a function $f$ that reduces the metric by a constant factor:
$$d(f(X), f(Y)) \leq \beta d(X, Y)$$
The factor $\beta\in[0,1)$ is called the \emph{modulus}. 
If $\beta=1$ then $f$ is called a \emph{nonexpansion}.
For a linear operator $A$, with metric $d(x, y) = \|x-y\|_P$, the modulus is the same as the operator norm $\|A\|_{P,P}$.
The \emph{Banach fixed-point theorem} guarantees the existence of a fixed point of any contraction on a complete metric space.

\subsection{Norms for POMDPs and PSRs}

We can bound the transition operators $T_{ao}$ for POMDPs and PSRs using operator norms that correspond to the set of valid states. 
In POMDPs, valid belief states are probability distributions, and therefore satisfy $\|q\|_1\leq 1$.
For PSRs, there is no single norm that works for all models. Instead, for each PSR, we only know that there exists a norm $\|\cdot\|_{\bar S}$ such that all valid states are in the unit ball $\bar S$. (We can get $\bar S$ by symmetrizing the PSR's set of valid states $S$.)
We will write $\|\cdot\|_{\bar S}$ in both cases, by taking $S$ to be the probability simplex if our model is a POMDP\@.
Given these definitions, we are guaranteed that, for each $a$,
$$\|T_a\|_{S} \leq 1 \quad\mathrm{where}\quad T_a = \sum_o T_{ao}$$
We also know that each transition operator $T_{ao}$ maps states to unnormalized states: it maps $S$ to the cone generated by $S$, i.e., $\{\lambda s \mid s\in S, \lambda \geq 0 \}$.

\subsection{Convergence: key step}

The key step in the proof of convergence is to analyze
$$\sum_o \Phi T_{ao}$$
for a fixed action $a$. We will show that this operation is a nonexpansion in the Hausdorff metric based on a particular norm.
To build the appropriate norm, we can start from norms for our states and our features. 
For states we will use the norm that corresponds to our state space: $\|\cdot\|_{\bar S}$.
For features we can use any norm $\|\cdot\|_F$.
For elements of $\Phi$ we can then use the operator norm for $\bar S$ and $F$: $\|\cdot\|_{F,\bar S}$. 
For sets like $\Phi$ we can use the Hausdorff metric based on $\|\cdot\|_{F,\bar S}$, which we will write as just $d(\cdot, \cdot)$.

For simplicity we will first analyze distance to a point:
start by assuming $d(\Phi,\{0\}) \leq k$ for some $k$. 
Now, for each $a$,
\begin{align*}
\textstyle
d(\sum_o \Phi T_{ao}, \{0\})
&= \textstyle
\sup_{\psi_o\in\Phi T_{ao}} \|\sum_o \psi_o\|_{F,\bar S} \\
&= \textstyle
\sup_{\phi_o\in\Phi} \|\sum_o \phi_o T_{ao}\|_{F,\bar S} \\
&= \textstyle
\sup_{\phi_o\in\Phi}\sup_{f\in F^*,q\in \bar S} f^T\sum_o \phi_o T_{ao}q
\end{align*}
where we have written $\sup_{\psi_o\in\Phi T_{ao}}$ as shorthand for $\sup_{\psi_1 \in \Phi T_{a,1}} \sup_{\psi_2\in\Phi T_{a, 2}} \ldots$, i.e., one supremum per observation.

Since $q$ is the solution to a linear optimization problem, we can assume it is an extreme point of the feasible region $\bar S$, which means either $q\in S$ or $q\in -S$. Assume $q\in S$; the other case is symmetric. This lets us replace $\sup_{q\in\bar S}$ with $\sup_{q\in S}$.

We next want to simplify the supremum over $f$. We can do this in two steps: first, the supremum can only increase if we let the choice of $f$ depend on $o$ (which we write as $\sup_{f_o}$). Second, H\"older's inequality tells us that $\|\phi_o^T f_o\|_{\bar S^*}\leq k$, since $\|f_o\|_{F^*}\leq 1$ and $\|\phi_o^T\|_{\bar S^*, F^*}\leq k$. So, optimizing over $k \bar S^*$ instead of just over vectors of the form $\phi_o^T f_o$ can again only increase the supremum. We therefore have
\begin{align*}
\lefteqn{\textstyle d(\sum_o \Phi T_{ao}, \{0\})} \quad & \\
&\leq \textstyle
\sup_{\phi_o\in\Phi}\sup_{q\in S, f_o\in F^*} \sum_o f_o^T \phi_o T_{ao} q\\
&\leq \textstyle
\sup_{q\in S} \sup_{r_o\in k \bar S^*} \sum_o r_o^T T_{ao} q
\end{align*}
We can now solve the optimizations over $r_o$. Note that the normalization vector $u$ is in $\bar S^*$: $u\cdot s = 1$ for every $s\in S$, so $u\cdot \bar s\in [-1,1]$ for every $\bar s\in\bar S$. And, for any valid state $s$, no vector in $\bar S^*$ can have dot product larger than $1$ with $s$, by definition of $\bar S^*$. $T_{ao}q$ is a nonnegative multiple of a valid state for each $o$; therefore, $r_o=ku$ is an optimal solution for each $o$, and we have
\begin{align*}
{\textstyle d(\sum_o \Phi T_{ao}, \{0\})} 
&\leq \textstyle
\sup_{q\in S} \sum_o ku^T T_{ao} q \\
&= \textstyle k\, \sup_{q\in S} u^T T_a q \\
&= k = d(\Phi, \{0\})
\end{align*}
To handle distances to a general set $\Phi$, we need to track a $\sup\inf$ instead of just a $\sup$. 
Assume wlog that 
$$\textstyle d(\sum_o \Phi T_{ao},\sum_o \Psi T_{ao}) = \bar d(\sum_o \Phi T_{ao},\sum_o \Psi T_{ao})$$
(the other ordering is symmetric). Then
\begin{align*}
    \lefteqn{\textstyle \bar d(\sum_o \Phi T_{ao},\sum_o \Psi T_{ao})} \quad & \\
    &= \textstyle \sup_{\phi_o\in\Phi} \inf_{\psi_o\in\Psi} \|\sum_o \phi_o T_{ao} -\sum \psi_o T_{ao}\|_{F,\bar S} \\
    &= \textstyle \sup_{\phi_o\in\Phi} \inf_{\psi_o\in\Psi} \|\sum_o (\phi_o-\psi_o) T_{ao} \|_{F,\bar S}
\end{align*}
The argument proceeds from here exactly as above, since we know that $\|\phi_o-\psi_o\|_{F,\bar S}$ is bounded by $d(\Phi,\Psi)$ for each $o$.

\subsection{Convergence: rest of the proof}

The remaining steps in our dynamic programming update are multiplying by $\gamma$, adding $F_a$, and taking the convex hull of the union over $a$. Multiplying the sets by $\gamma$ changes the modulus from $1$ to $\gamma$. Adding the same vector to both sets does not change the modulus. Finally, convex hull of union also leaves the modulus unchanged: more specifically, if $f_1, f_2, \ldots$ are all contractions of modulus $\gamma$, then the mapping
$$\Phi \to \mathrm{conv}\bigcup_i f_i(\Phi)$$
is also a contraction of modulus $\gamma$. To see why, consider two sets $\mathrm{conv}\cup_i f_i(\Phi)$ and $\mathrm{conv}\cup_i f_i(\Psi)$, with $d(\Phi,\Psi)=1$. Consider a point in the former set: it can be written as $\sum_j \alpha_j \phi_j$ with each $\phi_j$ in one of the sets $f_i(\Phi)$ and the $\alpha_j$ a convex combination. For each $j$, we can find a point in the corresponding set $f_i(\Psi)$ at distance at most $\gamma$, since $f_i$ is a contraction. Using the triangle inequality on the convex combination, the final distance is therefore at most $\gamma$.

Putting everything together, we have that the dynamic programming update is a contraction of modulus $\gamma < 1$. From here, the Banach fixed-point theorem guarantees that there exists a unique fixed point of the update, and that each iteration of dynamic programming brings us closer to this fixed point by a factor $\gamma$, as long as we initialize with a nonempty compact subset of the set of matrices.

\section{Background on PSRs}
\label{sec:PSR-extra}

Here we describe a mechanical way to define a valid PSR, given some information about a controlled dynamical system. 
This method is fully general: if it is possible to express a dynamical system as a PSR, we can use this method to do so. And, PSRs constructed this way allow a nice interpretation of the otherwise-opaque PSR state vector. To describe this method, it will help to define a kind of experiment called a \emph{test}.

\subsection{Tests}

A test $\tau$ consists of a sequence of actions $A_\tau = (a_1, a_2, \ldots, a_\ell)$ and a function $F_\tau:\{1\ldots O\}^\ell\to\mathbb R$. We execute $\tau$ by executing $a_1, a_2, \ldots, a_\ell$ starting from some state $q$. We record the resulting observations $o_{t}, o_{t+1}, \ldots, o_{t+\ell-1}$, and feed them as inputs to $F_\tau$; the output is called the \emph{test outcome}.
The \emph{test value} is the expected outcome
$$\tau(q) = \mathbb E(F(o_{t}, o_{t+1}, \ldots, o_{t+\ell-1}) \mid q_t=q, \mathrm{do}~A^\tau)$$
A \emph{simple test} is one where the function $F_\tau$ is the indicator of a given sequence of $\ell$ observations; in this case the test value is also called the test \emph{success probability}. Tests that are not simple are \emph{compound}. Below, we will use tests to construct PSRs. If we use exclusively simple tests, we will call the result a \emph{simple PSR}; else it will be a \emph{transformed PSR}.

We can express compound tests as linear combinations of simple tests: we can break the expectation into a sum over all possible sequences of $\ell$ observations to get 
$$\tau(q) = \sum_{o_1\ldots o_\ell} P(o_1\cdots o_\ell\mid q, \mathrm{do}~A^\tau)\, F^\tau(o_1, \ldots, o_\ell)$$
and each term in the summation is a fixed multiple of a simple test probability.

In a PSR, for any test $\tau$, it turns out that the function $\tau(q)$ is \emph{linear}: for a simple test with actions $a_1\ldots a_\ell$ and observations $o_1\ldots o_\ell$, 
\begin{eqnarray*}
\tau(q) &=& P(o_1,\ldots,o_\ell\mid q, \mathrm{do}~A^\tau) \\
&=& u^T T_{a_\ell o_\ell}\cdot T_{a_{\ell-1}o_{\ell-1}} \cdots T_{a_2o_2}\cdot T_{a_1o_1} q
\end{eqnarray*}
which is linear in $q$. For a compound test, the value is linear because it is a linear combination of simple tests.

In fact, this linearity property is the defining feature of PSRs: a dynamical system can be described as a PSR exactly when we can define a state vector that makes all test values into linear functions. That is, we can write down a PSR iff there exist state extraction functions $q_t = Q_t(a_1, o_2, a_2, o_2, \ldots, a_{t-1}, o_{t-1})\in\mathbb R^k$ such that, for all tests $\tau$, there exist prediction vectors $m_\tau\in\mathbb R^k$ such that the value of $\tau$ is $\tau(q_t)=m_\tau\cdot q_t$. 
There may be many ways to define a state vector for a given dynamical system; we are interested particularly in \emph{minimal} state vectors, i.e., those with the smallest possible dimension $k$.

Above, we saw one direction of the equivalence between PSRs and dynamical systems satisfying the linearity property: given a PSR, the state update equations define $Q_t$, and the expression above gives $m_\tau$. We will demonstrate the other direction in the next section below, by constructing a PSR given $Q_t$ and $m_\tau$.

Given a test $\tau$, an action $a$, and an observation $o$, define the \emph{one-step extension} $\tau^{ao}$ as follows: let $a_1,\ldots,a_\ell$ be the sequence of actions for $\tau$, and let $F(\cdot)$ be the statistic for $\tau$. Then the action sequence for $\tau^{ao}$ is $a, a_1, \ldots, a_\ell$, and the statistic for $\tau^{ao}$ is $F^o(\cdot)$, defined as
$$F^o(o_1,\ldots,o_{\ell+1}) = {\mathbb I}(o_1 = o) F(o_2,\ldots, o_{\ell+1})$$
In words, the one-step extension tacks $a$ onto the beginning of the action sequence. It then applies $F(\cdot)$ on the observation sequence starting at the \emph{second} time step in the future, but it either keeps the result or zeros it out, depending on the value of the first observation.

We can relate the value of a one-step extension test $\tau^{ao}$ to the value of the original test $\tau$:
$$\tau^{ao}(q) = P(o\mid q, \mathrm{do}~a) \tau(q')$$
where $q'=T_{ao}q/u^TT_{ao}q$ is the state we reach from $q$ after executing $a$ and observing $o$.
(We can derive this expression by conditioning on whether we receive $o$ or not: with probability $P(o\mid q, \mathbf{do}~a)$ the outcome of $\tau^{ao}$ is as if we executed $\tau$ from $q'$, else the outcome of $\tau^{ao}$ is zero.)

For example, in any PSR, we can define the \emph{constant test} $\tau_{\mathbf 1}$, which has an empty action sequence and always has outcome equal to $1$. The one-step extensions of this test give the probabilities of different observations at the current time step: 
$$\tau_{\mathbf 1}^{ao}(q) = P(o\mid q, \mathrm{do}~a)$$

\subsection{PSRs and tests}
We can use tests to construct a PSR from a dynamical system, and to interpret the resulting state vector. This interpretation explains the terminology \emph{predictive state}: our state is equivalent to a vector of predictions about the future. Crucially, these predictions are for observable outcomes of experiments that we could actually conduct. This is in contrast to a POMDP's state, which may be only partially observable. 

In more detail, suppose we have a dynamical system with a minimal state $q_t$ that satisfies the linearity property defined above. That is, suppose we have functions $Q_t$ that compute minimal states $q_t = Q_t(a_1, o_1,\ldots, a_{t-1},o_{t-1})\in\mathbb R^k$, and vectors $m_\tau\in\mathbb R^k$ that predict test values $\tau(q_t)=m_\tau\cdot q_t$. We will show that each coordinate of $q_t$ is a linear combination of test values, and we will define PSR parameters $T_{ao}, u$ that let us update $q_t$ recursively, instead of having to compute $q_t$ from scratch at each time step using the state extraction functions $Q_t$.

Pick $k$ tests $\tau_1\ldots\tau_k$, and define $q'_t\in\mathbb R^k$ to have coordinates $[q'_t]_i = m_{\tau_i}\cdot q_t$. Equivalently, let $S$ be the matrix with rows $m_{\tau_i}$, and write $q'_t=Sq_t$.
We say that our set of tests is linearly independent if their prediction vectors $m_{\tau_i}$ are linearly independent \textemdash\ equivalently, if the matrix $S$ is invertible.
If this happens to be true for $\tau_1\ldots\tau_k$, then $q_t'$ is another minimal state vector for our dynamical system: the value of test $\tau$ is $m_\tau\cdot q_t = m_\tau\cdot S^{-1}q'_t$, which is a linear function of $q'_t$. Furthermore, we have interpreted each coordinate of $q_t$ as a linear combination of tests, as promised: $q_t = S^{-1}q'_t$.

It turns out that we can always pick $k$ linearly independent tests. To see why: the empty list is linearly independent. For any list shorter than $k$, there will always exist another linearly independent test that we can add: if not, every possible $m_\tau$ is a linear combination of our existing vectors $m_{\tau_i}$, meaning that we can express $m_\tau \cdot q_t$ as a linear function of $m_{\tau_i}\cdot q_t$. We could then define $[q'_t]_i=m_{\tau_i}\cdot q_t$ as before, and get a state vector of dimension smaller than $k$, contradicting the minimality of $q_t$.

Now it just remains to show how to update our state vector recursively. We will describe first how to update $q'_t$, and then how to update the original state vector $q_t$.

For each of the tests $\tau_i$ that make up $q_t'$, consider the one-step extensions $\tau_i^{ao}$ for each $a$ and $o$. Write $m_i^{ao}$ for the corresponding prediction vectors, so that $\tau^{ao}_i(q_t') = m_i^{ao}\cdot q_t'$. And, write $m_{\mathbf 1}$ for the prediction vector of the constant test $\tau_{\mathbf 1}$.

We can now define PSR parameters in terms of these prediction vectors: let $T_{ao}$ be the matrix with rows $m_i^{ao}$,
$$[T_{ao}]_{ij} = [m_i^{ao}]_j$$
and define 
$$u = m_{\mathbf 1}$$
If we now use $T_{ao}$ to update $q_t'$, we get
\begin{eqnarray*}
[T_{ao} q'_t]_i &=& m_i^{ao}\cdot q'_t\\
&=& \tau_i^{ao}(q'_t)\\
&=& P(o\mid q'_t, \mathrm{do}~a) \tau_i(q'_{t+1})\\
&=& P(o\mid q'_t, \mathrm{do}~a) [q'_{t+1}]_i
\end{eqnarray*}
or equivalently
$$q'_{t+1} = T_{ao} q'_t\, /\, P(o\mid q'_t, \mathrm{do}~a)$$
which is the correct update for $q'_t$ after action $a$ and observation $o$. And,
\begin{eqnarray*}
u\cdot T_{ao} q'_t &=& u\cdot q'_{t+1}\, P(o\mid q'_t, \mathrm{do}~a)\\
&=& m_{\mathbf 1}\cdot q'_{t+1}\, P(o\mid q'_t, \mathrm{do}~a)\\
&=& P(o\mid q'_t, \mathrm{do}~a)
\end{eqnarray*}
demonstrating that $u$ correctly computes observation probabilities and lets us normalize our state vector.

Recapping, if we use the new state vector $q_t'$, each coordinate of our state is a test value, and we can interpret our parameter matrices in terms of tests. The rows of $T_{ao}$ correspond to one-step extension tests, and the normalization vector $u$ corresponds to the constant test.

For the original state vector $q_t$, we can make a similar interpretation. Define the one-step extension of a linear combination of tests by passing the extension through the linear combination: that is, given a linear combination $\sigma = \sum_i a_i \tau_i$ for coefficients $a_i$ and tests $\tau_i$, the one-step extension $\sigma^{ao}$ is $\sum_i a_i \tau_i^{ao}$. With this definition, the exact same construction of $T_{ao}$ and $u$ works for our original state vector. That is, each component of $q_t$ can be interpreted as a linear combination of tests; each row of $T_{ao}$ is the prediction vector for a one-step extension of one of these linear combinations; and $u$ is the prediction vector for the constant test.

\section{Background on policies}
\label{sec:policy-extra}

We can represent a horizon-$H$ deterministic policy $\pi$ as a balanced tree of depth $H$ (Fig.~\ref{fig:policy_tree}).
We start at the root of the tree. At each node, we execute the corresponding action $a$, branch to a child node $\pi(o)$ depending on the resulting observation $o$, and repeat. 

We can write a horizon-$H$ stochastic policy as a mixture of horizon-$H$ deterministic policies \textemdash\ i.e., a convex combination of depth-$H$ trees. To execute a stochastic policy, we alternate between choosing actions and receiving observations, as follows.
To choose an action, we look at the labels of the root nodes of all of the policy trees in our mixture: the probability of action $a$ is the total weight of trees whose root label is $a$. Given the action $a$, we keep only the trees with root label $a$, and renormalize the mixture weights to sum to 1. 
To incorporate an observation, we branch to a child node within each tree according to the received observation. That is, we replace each tree $\pi$ in our mixture by its child $\pi(o)$, keeping the same weight. We write $\pi(a,o)$ for the resulting mixture after choosing action $a$ and incorporating observation $o$.

\section{Successor feature matrices}
\label{sec:succ-mat}

In a POMDP or PSR, we do not want a separate successor feature vector at each state, since we do not have access to a fully observable state. Instead, the successor feature representation is a function of the continuous predictive state or belief state $q$. Here, we show that this function is \emph{linear} in $q$: that is, we can represent it as 
$$\phi^\pi(q) = A^\pi q$$
for some matrix $A^\pi\in\mathbb R^{d\times k}$ that depends on the policy $\pi$. We also show how to compute the {successor feature matrix} $A^\pi$.

We can show linearity, and at the same time compute $A^\pi$, by induction over the horizon. In the base case (a horizon of $H=1$), our total discounted features are the same as our one-step features:
$$\phi^\pi(q) = \mathbb E_\pi[f(q, a)] = \sum_a P(a\mid\pi) F_a q$$
Note that the RHS is a linear function of $q$, as claimed. 

In the inductive case (horizon $H>1$), we can split our our expected total discounted features into contributions from the present and the future:
$$\phi^\pi(q_t) = \mathbb E_\pi[f(q_t, a_t) + \gamma \phi^{\pi(a_t,o_t)}(q_{t+1})]$$
In the the contribution of future time steps, note both the one-step updated policy $\pi(a_t,o_t)$ and the one-step updated predictive state $q_{t+1}$.
Expanding the expectation and substituting our expression for $f(\ldots)$, we get
$$\phi^\pi(q_t) = \sum_{a,o} P(a\mid \pi) P(o\mid q_t, \mathrm{do}~a) [F_a q_t + \gamma \phi^{\pi(a,o)}(q_{t+1})]$$
We can inductively assume that $\phi^{\pi(a,o)}$ is linear, since $\pi(a,o)$ is a shorter-horizon policy than $\pi$. That is, we can write $\phi^{\pi(a,o)}(q) = A^{\pi(a,o)}q$. Substituting this expression, and using
$q_{t+1} = T_{ao} q_t \,/\, P(o \mid q_t, \mathrm{do}~a)$,
we see that $P(o\mid q_t, \mathrm{do}~a)$ cancels:
$$\phi^\pi(q_t) = \sum_a P(a\mid \pi) \left[F_a q_t + \gamma \sum_o A^{\pi(a,o)}T_{ao} q_t\right]$$
We can observe that the RHS is a linear function of $q_t$, which completes our inductive proof of linearity.

Because of linearity, there exists a matrix $A^\pi$ such that $\phi^\pi(q) = A^\pi q$. With this notation,
$$A^\pi q_t = \sum_a P(a\mid \pi) \left[F_a q_t + \gamma \sum_o A^{\pi(a,o)}T_{ao} q_t\right]$$
Because the above equation must hold for any predictive state $q_t$, we get
$$A^\pi = \sum_a P(a\mid \pi) \left[F_a + \gamma \sum_o A^{\pi(a,o)}T_{ao} \right]$$
This equation defines $A^\pi$ recursively in terms of matrices for shorter-horizon policies. 
So, we can compute $A^\pi$ by dynamic programming, working backward from horizon 1: we start by computing the matrices for all 1-step policies that we can get from $\pi$ by fixing the first $H-1$ actions and observations, then combine these to compute the matrices for all 2-step policies that we can get from $\pi$ by fixing the first $H-2$ actions and observations, and so forth. 

For a deterministic policy with root action $a$, the recursion simplifies to
$$A^\pi = F_a + \gamma \sum_o A^{\pi(o)} T_{ao}$$
We can think of this recursion as working upward from the leaves of a single policy tree.

\section{Implementation and experimental setup}
In the following sections we discuss experimental details for computing the successor feature sets and using them for feature matching.
\label{sec:implementation}

\subsection{Successor Feature Sets Implementation}

We start by initializing each $\Phi_{ao}$ to the set consisting of the zero matrix with dimension $d\times k$. We sample a fixed set of directions $m_i\in\mathbb R^{d\times k}$ in a $d k$-sphere by sampling from a Gaussian and normalizing. To make computation more regular and GPU-friendly, we pre-allocate $|A|$ tensors whose dimensions are $\hat{m} \times d \times k$; we group the $\Phi_{ao}$ matrices for all $o$ and store them into the tensors. $\hat{m}$ corresponds to the max number of boundary points that we store for each $\Phi_{ao}$. These tensors allow us efficiently solve  
$$\textstyle \arg\max\, \langle m_i , \phi\rangle \mbox{ for } \phi \in \bigcup_{a'} \left[ F_{a'} + \gamma\sum_{o'} \Phi_{a'o'}\right] T_{ao} $$
because $\left[ F_{a'} + \gamma\sum_{o'} \Phi_{a'o'}\right] T_{ao}$ becomes a series of matrix multiplications which we can efficiently compute in parallel using a GPU. We try three different numbers of random projections: 50, 100 and 175. We prune the resulting boundary points to keep only the unique ones.

\subsection{Mountain-Car Implementation}
In the mountain-car environment, the one-step features are radial basis functions of the state with values in $[0,1]$. In particular, if we rescale the state space to $[-1,1]\times[-1,1]$, we set the 9 RBF centers to be at $\{-0.8,0,0.8\} \times \{-0.8,0,0.8\}$, a $3 \times 3$ grid. The RBF widths in the rescaled state space are $\sigma = 0.8$.

\subsection{Grid POMDP Implementation}
In the Grid POMDP environment the agent has 0.05 probability of transitioning to a random neighboring state, and an 0.05 probability of observing a random neighboring state instead of the current state that it is in. We experimented as well with various amounts of noise (not shown); increasing the noise increases the effective dimensionality of the $\Phi_{ao}$ sets, and we start to need more and more boundary points. Decreasing the noise makes the POMDP solution approach the MDP solution.

\subsection{Feature Matching Implementation}
To implement step 5 or step 7 in Algorithm~\ref{alg-featmat}, we need to solve a small convex program. The best way to do so depends on the data structures we use to represent $\Phi$ and $\Phi_a$. With our $\Phi_{ao}$ decomposition, the sets $\Phi_a$ are the convex hull of a finite set of vertices, with the number of vertices bounded by $\hat m^{|O|}$.

With this representation, for step 5, a reasonable approach is to use the Frank-Wolfe algorithm to find $\phi_{it}$ and $p_{it}$: if we minimize the squared error between the LHS and RHS of the equation in step 5, the Frank-Wolfe method will naturally produce its output in the form of a convex combination of vertices of $\Phi_a q_t$.

Note that if we use Frank-Wolfe in step 5, every $\phi_{it}$ we need to decompose in step 7 will be a vertex of one of the sets $\Phi_a q_t$. So, a reasonable approach is to annotate the vertices of $\Phi_a$ as we compute them. Each vertex of $\Phi_a$ will be constructed from some list of vertices of $\Phi_{ao}$ for different $o$'s; we can just record which vertices of the sets $\Phi_{ao}$ were used to construct each vertex of $\Phi_a$. We can multiply the current state $q_t$ into the vertices of $\Phi_a$ and $\Phi_{ao}$ to get vertices of $\Phi_a q_t$ and $\Phi_{ao} q_t$.

Note that if we use the above approach to decompose $\phi_{it}$, on the next time step $\phi^d_{t+1}$ will be a vertex of $\Phi q_{t+1}$. So, we will not need to run Frank-Wolfe in step 5 on any subsequent time steps, unless numerical errors or incomplete convergence of the dynamic programming iteration cause us to drift away from being an exact vertex.

\section{Convergence of point-based approximations}

\begin{figure}
    \centering
    \includegraphics[width=0.8\linewidth]{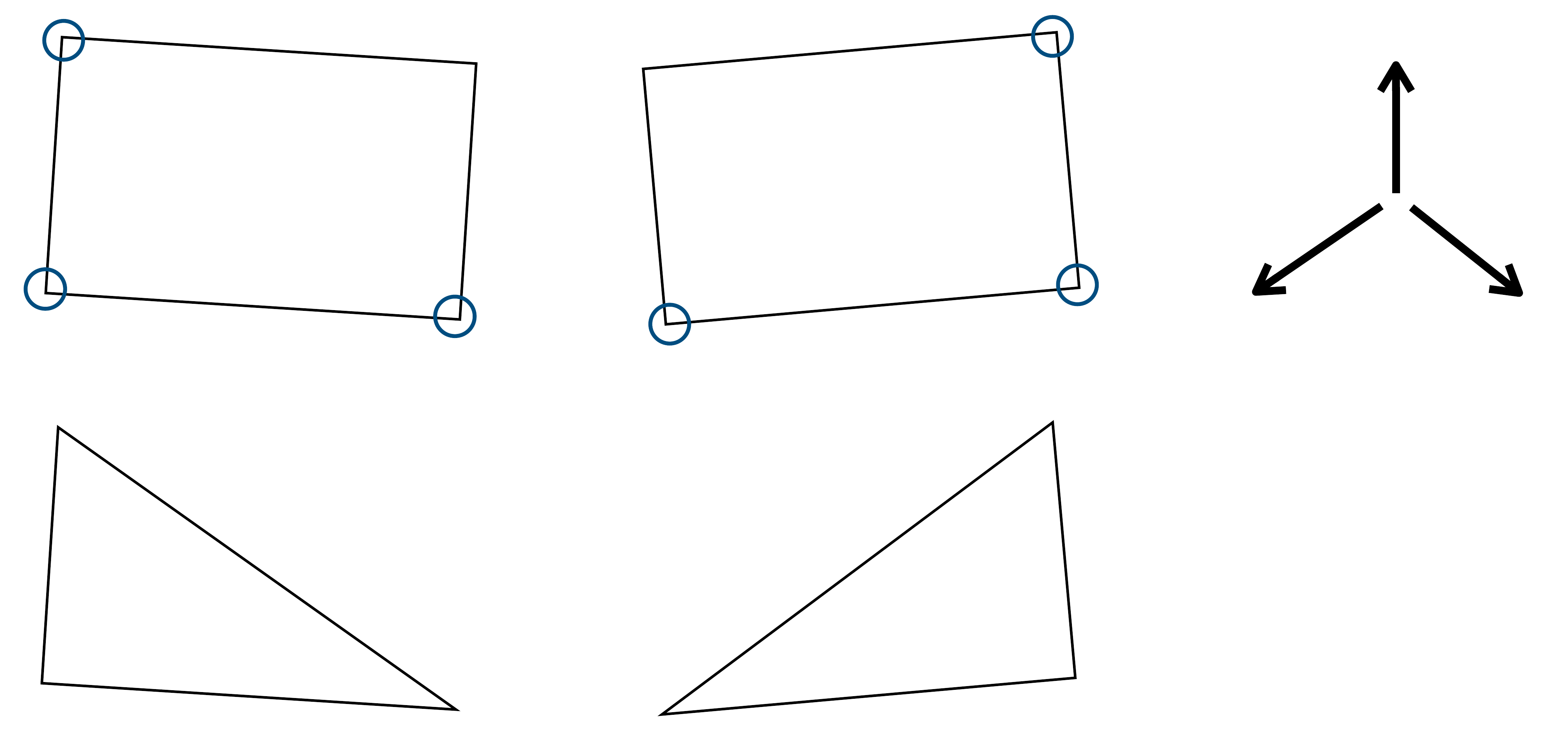}
    \caption{Example of the behavior of point-based approximation. Two convex sets (top row) are very similar. If we retain the maximal points (blue circles) in the indicated directions (arrows, top right), the convex hulls of the two sets of retained points are very different (bottom row).}
    \label{fig:point_based}
\end{figure}

While the exact dynamic programming update is a contraction, the point-based approximate dynamic programming update might not be. Fig.~\ref{fig:point_based} shows why: an arbitrarily small change in a backed up set can lead to a large change in the point-based approximation of that set. Despite this fact, in practice we observe rapid convergence of the point-based approximate iteration.

Nonetheless, we can show that a small modification of our point-based approximate method converges and has bounded error. In particular, we analyze \emph{monotone} point-based backups. Our analysis is similar to a corresponding analysis for monotone point-based value iteration in POMDPs.

For the modification, suppose that we are in a \emph{stoppable} process: that is, suppose there is a designated \emph{stop} action that ends the process, giving us some (possibly bad) terminal reward that can depend on the current state. In this case we can initialize our dynamic programming iteration with $\{\phi^{\rm stop}\}$, the singleton set containing the successor feature matrix of the policy that always takes the stop action. 
One common way that stoppable processes arise is if we have an \emph{emergency} or \emph{safety} policy --- the equivalent of a big red button that causes our robot to shut down or retreat to a safe state. If we have an \emph{idle} action, one that does not change our state but also does not yield a good reward, then we can use the always-idle policy as our safety policy.

In stoppable problems, with the given initialization, we know that our point-based backup will compute only \emph{achievable} successor feature matrices --- i.e., only those $\phi$ that correspond to policies that we can always execute. So, we can use monotone backups: we can keep at each step the better of the existing (horizon $H$) and the backed up (horizon $H+1$) successor feature matrix in each direction. (We can make the same modification to the exact backup operator as well: we merge together the current successor feature set $\Phi^{(H)}$ with the backed up successor feature set $\Phi^{(H+1)}$, by taking the convex hull of their union. This modification does not affect the convergence proof or error bound given above.)

We can now analyze the monotone backup.
First, note that the point-based backup of any set is a subset of the exact backup of that set, since we get the point-based backup by dropping elements of the exact backup.
Second, note that both the point-based and the exact backup operators are monotone with respect to set inclusion: if $P\subseteq Q$ then the backup of $P$ is a subset of the backup of $Q$.
So, the iterates from either are monotonically increasing. For the point-based backup, this means that the convex hull of our retained $\phi$ matrices at each horizon always contains the convex hull at shorter horizons. 

The exact backup sequence converges to the exact successor feature set, which is therefore an upper bound on the approximate backup sequence. By the monotone convergence theorem, this means that the monotone point-based iteration must converge to a subset of the exact successor feature set. 

We can use this same argument to get a simple error bound: write $\Phi^{PB}$ for the convergence point of the point-based iteration, and write $\Phi^{PB+}$ for its one-step exact backup. Suppose that these two sets differ by at most $\epsilon$ in Hausdorff metric. Then a standard argument shows that $\Phi^{PB}$ cannot be farther than $\frac{\epsilon}{1-\gamma}$ from the exact successor feature set. We know that $\epsilon$ can be at most the size of the exact successor feature set (which is bounded by $\frac{R_{\rm max}}{1-\gamma}$), but it may be much smaller.

\end{document}